\documentclass[11pt]{article}

\usepackage[final]{acl}

\usepackage{times}
\usepackage{latexsym}
\usepackage{tikz}
\usepackage{longtable}

\usepackage[T1]{fontenc}

\usepackage[utf8]{inputenc}

\usepackage{microtype}

\usepackage{inconsolata}

\usepackage{graphicx}
\usepackage{booktabs}

\title{What Carries Forward? Pretraining Disclosure Across Language Model Releases}

\author{Md Sakib Ul Islam\thanks{Equal contribution.} \\
  Victorian Institute of Technology, Australia \\
 \\\And
  S M Mehedi Zaman\footnotemark[1] \\
  Rutgers University, USA \\
  }

\begin{document}
\maketitle
\begin{abstract}
Understanding how language models were pretrained
requires information about their data, processing
procedures, and computational resources. However,
it remains unclear whether successive releases
preserve or expand earlier disclosure.
We conduct a document-based audit of 41 model entries
from 17 families using 24 pretraining-related indicators
from the 2025 Foundation Model Transparency Index (FMTI).
We examine overall scores and indicator profiles,
tracking gains, losses, and retained disclosures
across 24 adjacent comparisons in ten families.
Scores decrease in 11 comparisons, increase in six,
and remain unchanged in seven, with 38 indicator
losses and 28 gains overall.
Stable or improving scores can conceal losses:
OLMo~2 and OLMo~3 receive identical totals despite
three gains and three losses, while Llama~3.1 improves
over Llama~2 while losing credit for two data-related
criteria. Gaps also persist in corpus composition
and training compute.
These findings show that later releases do not
consistently preserve earlier pretraining disclosure
and motivate documentation that explicitly identifies
changes and the applicability of inherited information.
\end{abstract}

\section{Introduction}
\label{sec:introduction}

Pretraining choices affect what language models learn,
memorize, and represent. Corpus audits have identified
benchmark contamination, uneven multilingual data quality,
and filtering that disproportionately excludes text
associated with minority identities
\citep{dodge-etal-2021-documenting,
kreutzer-etal-2022-quality}.
Experiments show that deduplication reduces memorized
outputs and train--test overlap
\citep{lee2022deduplicating, deng-etal-2024-investigating}, while data age,
composition, and filtering affect performance,
with some effects persisting after fine-tuning
\citep{longpre-etal-2024-pretrainers}.
Training examples can also be recovered from model
outputs, including material containing personal
information \citep{carlini2021extracting}.
Information about pretraining is therefore necessary
for investigating model behavior and interpreting
evaluation results.

Yet access to model weights does not ensure access
to training data, code, or development documentation
\citep{liesenfeld2023opening,liesenfeld2024rethinking}.
Naming a source such as Common Crawl leaves unresolved
which material was selected and how it was transformed
\citep{baack2024commoncrawl,albalak2024selection}.
Documentation audits identify persistent gaps
\citep{oreamuno2024documentation,pepe2026datasets},
while insufficient train--test overlap reporting limits
the interpretation of benchmark results
\citep{zhang2025overlap}.
Descriptions of instruction tuning or alignment
provide complementary information; they cannot
substitute for an account of base-model pretraining.

We ask: \textit{How does pretraining disclosure change
across releases within model families?}
Existing studies examine developer-level transparency
and changes within an individual model family
\citep{bommasani2023fmti,wan20252025,xu2024tracing}.
We examine specific pretraining disclosures across
selected releases in multiple families.
Our focus is whether later releases preserve previously
documented information, and whether stable or improving
overall scores conceal losses in particular areas.

\begin{figure*}[t]
    \centering
    \includegraphics[width=\textwidth]
        {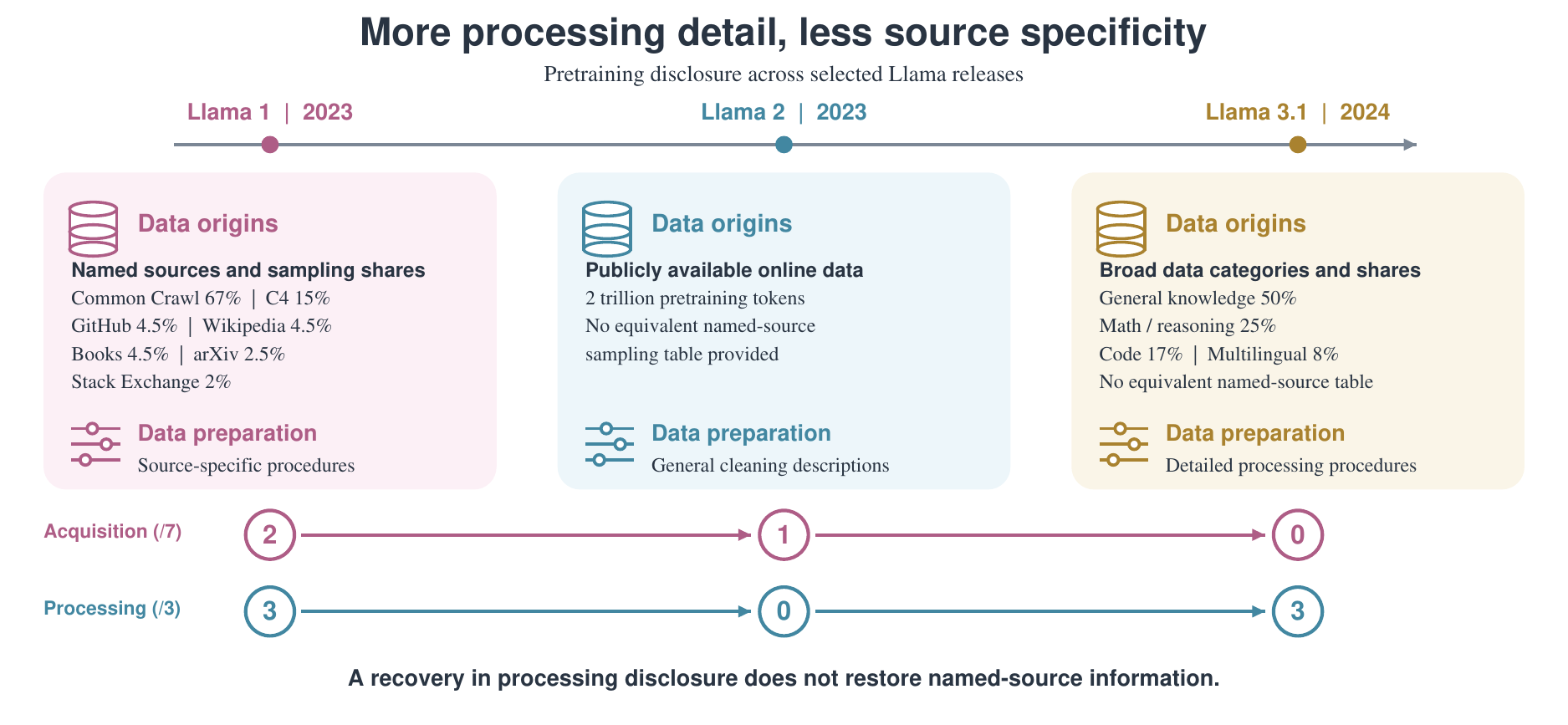}
    \caption{Pretraining disclosure across Llama
releases, based on their technical reports
\citep{touvron2023llama,touvron2023llama2,
dubey2024llama3}.
Later reports do not provide an equivalent to
Llama~1's named-source sampling breakdown, although
processing disclosure recovers in Llama~3.1.}
    \label{fig:disclosure-overview}
\end{figure*}

Figure~\ref{fig:disclosure-overview} illustrates this
distinction in the Llama family. Llama~1 reports
named data sources and sampling proportions.
Later reports provide broader descriptions or
category-level proportions without an equivalent
named-source breakdown. Processing disclosure recovers
in Llama~3.1, but acquisition scores continue to decline
under the selected criteria. More detailed documentation
of corpus preparation therefore need not provide
a more specific account of corpus origins.

We conduct a document-based audit of 41 model entries
from 17 families, comprising 27 open-weight and
14 open-source/reproducible entries.
Using a fixed subset of 24 pretraining-related indicators
from the 2025 FMTI, we examine technical reports and
associated official documentation. We compare overall
scores and indicator profiles, then trace gains,
losses, and retained disclosures across 24 adjacent
comparisons in ten families. Scores describe whether
the reviewed evidence meets the selected requirements,
rather than whether an undocumented practice occurred.

Later releases are not consistently better documented: scores decrease
in 11 comparisons, increase in six, and remain unchanged in seven,
with 38 indicator losses and 28 gains overall. Stable or improving totals
can conceal losses in particular information. These findings motivate
tracking both the extent and continuity of pretraining disclosure.

\section{Related Work}
\label{sec:related-work}

\subsection{Pretraining Data and Documentation}

Corpus audits identify properties that broad dataset descriptions leave unresolved.
\citet{dodge-etal-2021-documenting} find unexpected sources, benchmark contamination,
and disproportionate exclusion of minority-associated text in C4.
\citet{kreutzer-etal-2022-quality} identify quality problems and mislabeled
languages in multilingual web corpora. Controlled experiments show that
deduplication reduces memorization and overlap \citep{lee2022deduplicating},
while data age, composition, and filtering affect performance
\citep{longpre-etal-2024-pretrainers}. Surveys of selection methods and analyses
of Common Crawl further distinguish access to source data from knowledge of
how developers curate it \citep{albalak2024selection,baack2024commoncrawl}.
These studies motivate examining acquisition, processing, and corpus properties
separately. They investigate corpora or curation effects; we examine disclosure
about these aspects across model releases.

Pretraining also affects behavior addressed through later alignment.
\citet{minder2026synthetic} find effects of pretraining interventions under
identical subsequent post-training, although their approach also requires
later persona binding. Documentation of subsequent adaptation therefore
cannot substitute for an account of base-model development.

\subsection{Openness and Independent Scrutiny}

Openness assessments distinguish available weights from access to training
data, code, and technical documentation
\citep{liesenfeld2023opening,liesenfeld2024rethinking}.
The OSI distinction between open weights and Open Source AI informs our
operational grouping \citep{osiOpenWeights,osiOpenSourceAI}.
Higher disclosure in the open-source/reproducible group is partly expected,
since development-resource availability informs both classification and scoring.

Documentation audits identify reporting gaps in NLP papers and model repositories
\citep{magnusson-etal-2023-reproducibility,oreamuno2024documentation}.
Data Portraits support corpus-membership inspection
\citep{marone2023dataportraits}, while \citet{zhang2025overlap} advocate
training-data or overlap disclosure to support evaluation interpretation.
Our audit complements these approaches by assessing explicit requirements;
it does not inspect corpora, measure contamination, or test reproducibility.

\subsection{Transparency across Releases}

FMTI evaluates disclosure across upstream development, model information,
and downstream use \citep{bommasani2023fmti}, with transparency reports
consolidating supporting evidence \citep{bommasani2024reports}.
The 2024 assessment records improvements, whereas the 2025 edition identifies
renewed declines and persistent data and compute gaps
\citep{bommasani2024fmtiv11,wan20252025}. We apply a fixed pretraining subset
to releases rather than the full developer-level index.

\citet{xu2024tracing} trace OpenAI's GPT documentation, finding clearer
descriptions of limitations and downstream uses alongside more opaque
development information. \citet{pepe2026datasets} examine repository
documentation over time and identify persistent gaps concerning datasets,
bias, licensing, and terms of use. We extend these perspectives across
selected families, tracking gains, losses, and retained pretraining disclosures
under common criteria. This distinguishes changes in total scores from
changes in the information documented.

\section{Methodology}
\label{sec:methods}

We conduct a document-based audit of pretraining
disclosure for 41 selected model entries from
17 families. We examine base-model pretraining,
including continued pretraining where it forms
part of an included release. Disclosures concerning
instruction tuning, preference optimization,
reinforcement learning, and downstream fine-tuning
are excluded. We apply 24 selected indicators from
the 2025 Foundation Model Transparency Index
(FMTI; \citealp{wan20252025}) to compare documented
pretraining information across entries and within
model families.

\subsection{Model Selection}
\label{sec:model-selection}

We identified candidate models through web searches
and associated technical reports, developer websites,
and repositories. We included general-purpose text
language models and base-model releases available
for external use. Specialized models and instruction-tuned
or aligned derivatives were excluded unless their
documentation described a distinct qualifying
base-model pretraining release. For reports covering
multiple training stages, we considered only evidence
attributable to pretraining. Where a family included
multimodal models, we retained the text or base model
when it could be identified separately.

Inclusion normally required an official developer-authored
technical report or research paper. Marin~8B Base
and Marin~32B Base were exceptions: their detailed
official technical retrospectives were treated as
report-equivalent documentation.
We had a cutoff date for model-report search at 29 August 2026.
The corpus is a selected sample of qualifying entries,
rather than an exhaustive inventory of models available
by that date.

Each entry represents a release or a group of variants
documented together. For grouped entries, a criterion
received credit when the evidence met its requirements
for at least one included variant. The score therefore
summarizes disclosure available across the group
and does not imply equal documentation of every variant.

We distinguish \textit{Open Weights} ($n=27$) from
\textit{Open Source/Reproducible} ($n=14$), following
the distinction between weight availability and
broader access to development resources
\citep{osiOpenWeights,osiOpenSourceAI}.
The former provides model weights with limited
access to development artifacts; the latter provides
substantial access to training code and training-data
resources or information supporting external reproduction.


\subsection{Evidence Sources}
\label{sec:evidence-sources}

Each entry's qualifying technical report served as the primary
coding source. Directly associated official materials,
including model cards, repositories, dataset documentation,
and released training or data-processing code,
were consulted where they supplied evidence for
a selected indicator. These materials supplemented
the report. Third-party descriptions and assumptions
about undocumented practices were not used as evidence.

Evidence had to be explicitly attributable to the
included entry's pretraining process. Information
about post-training data, alignment procedures,
or downstream adaptation did not receive credit
as pretraining disclosure. Where information combined
training stages without establishing what applied
to pretraining, it was insufficient to support
the corresponding criterion.

For each decision, we recorded the disclosed information,
supporting source, scoring justification, and binary
score. The comparisons concern information documented
for each entry; they do not establish that earlier
documents were withdrawn.

We evaluated disclosure available for each release
at the time of assessment, rather than reconstructing
the documentation available on its original release date. Earlier documentation was considered when the later
release explicitly stated that the relevant data,
procedures, or other pretraining details remained
unchanged or were inherited. In those cases, we
consulted the earlier documentation for supporting
evidence. Without an explicit statement establishing
its applicability, earlier disclosure was not carried
forward to the later release.

\subsection{Disclosure Indicators}
\label{sec:disclosure-indicators}

We selected the indicator subset before model coding,
using the published FMTI definitions and their application
in the December 2025 transparency reports
\citep{stanfordcrfm2025fmtireports}.
We retained criteria concerning pretraining data,
acquisition, processing, corpus properties, compute,
development methods, model dependencies, and
train--test overlap. Criteria concerning deployment,
governance, and other topics outside this scope
were excluded.

The subset comprises seven acquisition indicators,
three processing indicators, five data-property
indicators, four compute indicators, three methods
indicators, and one indicator each for model dependencies
and train--test overlap.
Dependencies identify the development path underlying
the base model; overlap concerns the relationship
between pretraining data and evaluation datasets.

We retained the original indicator identifiers,
wording, definitions, notes, and binary requirements.
Their application was restricted to pretraining:
data criteria concerned the pretraining corpus,
compute criteria concerned pretraining resources,
and methods and code criteria concerned base-model
development. This scope restriction distinguishes
our assessment from the full FMTI.


\subsection{Coding and Verification}
\label{sec:coding}

ChatGPT (GPT-5.6, High reasoning setting) assisted
with initial evidence extraction and provisional coding.
Published FMTI transparency reports were supplied
as in-context examples of criterion application;
the model was not fine-tuned on these materials.
For each entry, it screened the qualifying documentation
and produced an evidence record, justification,
and provisional score.

Both authors subsequently verified the complete dataset.
One author checked every indicator against its sources
and the FMTI requirements. Another author then
checked all entries and indicators, including cases
not previously marked as uncertain. Discrepancies
were resolved by revisiting the sources and discussing
the relevant criterion until agreement was reached.
This procedure is similar to the worksheet verification
used by \citet{liu-etal-2024-ecbd}.
Because the second review examined existing coding
records rather than independently assigned labels,
we did not calculate inter-rater agreement.

An indicator received 1 when the qualifying evidence
met its requirements and 0 otherwise. A zero denotes
insufficient disclosure under the criterion, rather
than absence of the underlying practice. Explicit
statements of non-use were handled according to
the corresponding FMTI rules; non-use was not inferred
from silence. The equally weighted indicators were
summed to produce a pretraining disclosure score
out of 24 and descriptive subdomain totals.

We summarize scores and indicator prevalence and
compare adjacent entries in the selected family
sequences. Indicator transitions distinguish gains
($0\rightarrow1$), losses ($1\rightarrow0$), and
retained disclosures ($1\rightarrow1$).
The sequences include architectural branches,
continued-pretraining releases, and grouped variants;
these are descriptive comparisons rather than
controlled comparisons between equivalent successors.






\section{Results}
\label{sec:results}

\subsection{Distribution of Pretraining Disclosure}
\label{sec:disclosure-distribution}

Table~\ref{tab:disclosure-distribution} summarizes scores
for the 41 entries. The mean was 8.90 out of 24,
the median was 6, and scores ranged from 1 to 18.
No entry met all selected disclosure requirements.

\begin{table}[t]
\centering
\small
\setlength{\tabcolsep}{4pt}
\begin{tabular}{lrrrr}
\toprule
Group & $n$ & Mean & Median & Range \\
\midrule
Open weights & 27 & 5.19 & 5 & 1--10 \\
Open source/reproducible & 14 & 16.07 & 16 & 14--18 \\
\midrule
All entries & 41 & 8.90 & 6 & 1--18 \\
\bottomrule
\end{tabular}
\caption{Pretraining disclosure scores by operational
openness group. Scores count requirements met out of 24;
$n$ counts releases or grouped variants.}
\label{tab:disclosure-distribution}
\end{table}

Open-source/reproducible entries scored higher throughout
the selected corpus. Their lowest score, 14, exceeded
the highest open-weight score, 10. This separation
is partly expected: training-code and training-data
availability inform both the group definitions and
some disclosure criteria. The comparison characterizes
the selected entries and does not establish an
independent effect of openness on disclosure.

\subsection{Disclosure Changes across Releases}
\label{sec:disclosure-changes}

\begin{figure*}[t]
    \centering
    \includegraphics[width=\textwidth]
        {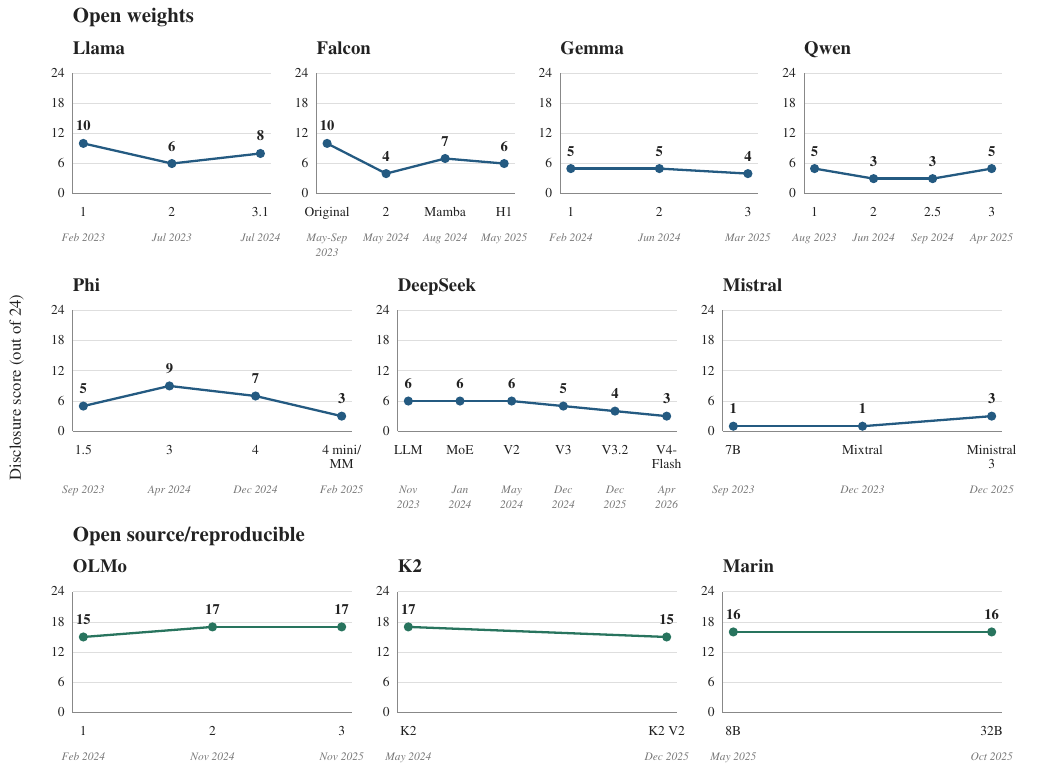}
    \caption{Pretraining disclosure scores across selected
    family sequences. Scores are out of 24.
    Release months and years appear below the version
    labels; the original Falcon group spans
    May--September 2023. Entries are equally spaced
    by order, not elapsed time. Connecting lines
    do not imply direct model derivation.
    MM denotes multimodal.}
    \label{fig:family-trajectories}
\end{figure*}

Later entries did not consistently meet more disclosure
requirements. Across 24 adjacent comparisons in ten
families, scores decreased in 11 cases, increased in six,
and remained unchanged in seven.
Figure~\ref{fig:family-trajectories} shows the selected
release sequences.

DeepSeek initially maintained a score of 6 across
three entries, followed by successive decreases
to 5, 4, and 3. Gemma scored 5, 5, and 4.
Llama and Falcon showed declines followed by partial
recoveries: Llama changed from 10 to 6 to 8,
and Falcon from 10 to 4 to 7 to 6.
Neither sequence returned to its initial score.

Other families followed different patterns.
Qwen returned to 5 after two entries scoring 3.
Phi increased from 5 to 9 before decreasing to 7
and then 3. Mistral increased from 1 for Mistral~7B
and Mixtral~8$\times$7B to 3 for Ministral~3.
Among open-source/reproducible entries, OLMo increased
from 15 to 17 and remained at 17, K2 decreased
from 17 to 15, and both Marin entries scored 16.

These patterns describe disclosure available for
the selected entries at assessment. The sequences
include architectural branches, continued-pretraining
releases, and grouped variants, rather than uniformly
comparable model generations.

\subsection{Aggregate Scores Conceal Changes in Disclosure}
\label{sec:disclosure-profile-changes}

Across the 24 comparisons, we recorded 28 indicator
gains ($0\rightarrow1$) and 38 losses ($1\rightarrow0$).
A loss means that the evidence applicable to the later
entry did not meet a previously satisfied requirement.
It does not establish that earlier documents were
withdrawn or that the underlying practice changed.

\begin{figure*}[t]
    \centering
    \includegraphics[width=\textwidth]
        {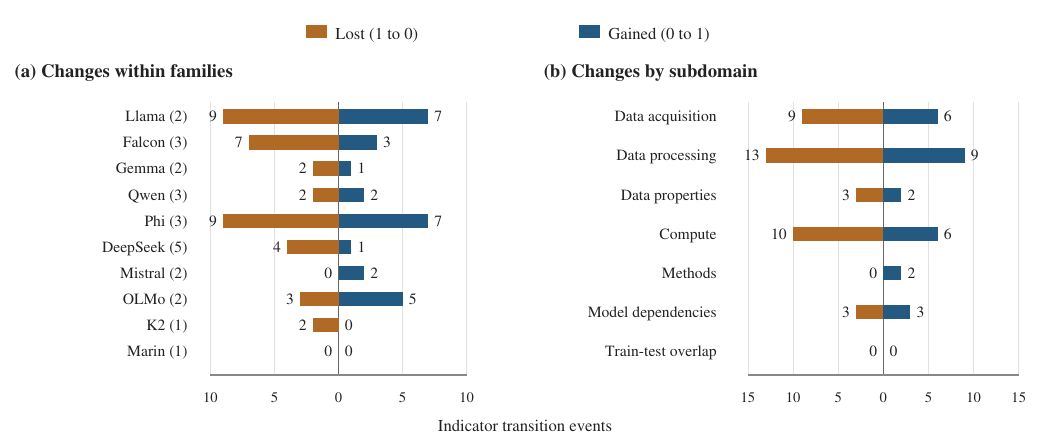}
    \caption{Indicator gains and losses across 24 adjacent
    comparisons, grouped by (a) family and (b) subdomain.
    Parentheses give the number of comparisons per family.
    Orange indicates losses ($1\rightarrow0$);
    blue indicates gains ($0\rightarrow1$).
    Counts include repeated transitions and are not
    normalized by comparisons or indicators.}
    \label{fig:disclosure-gains-losses}
\end{figure*}

Figure~\ref{fig:disclosure-gains-losses} summarizes
transitions by family and subdomain.
Llama and Phi each recorded seven gains and nine losses,
while Qwen recorded two gains and two losses.
Data processing accounted for nine gains and 13 losses;
compute accounted for six gains and ten losses.
These are transition counts, not rates: families
have different numbers of comparisons, and subdomains
contain different numbers of indicators.

Two illustrative comparisons show why totals alone
are insufficient (Table~\ref{tab:disclosure-profile-changes}).
OLMo~2 and OLMo~3 both scored 17, but six indicators
changed. OLMo~3 gained credit for crawling, data-domain
composition, and compute hardware, while losing credit
for final-training compute usage, compute provider,
and model dependencies. Fourteen criteria remained
satisfied. The unchanged total therefore reflected
three gains offset by three losses.

\begin{table}[t]
\centering
\small
\setlength{\tabcolsep}{4pt}
\begin{tabular}{lrrrr}
\toprule
Comparison & Scores & Gained & Lost & Retained \\
\midrule
OLMo 2 $\rightarrow$ 3
    & $17\rightarrow17$ & 3 & 3 & 14 \\
Llama 2 $\rightarrow$ 3.1
    & $6\rightarrow8$ & 4 & 2 & 4 \\
\bottomrule
\end{tabular}
\caption{Two illustrative release comparisons.
Gained, lost, and retained denote $0\rightarrow1$,
$1\rightarrow0$, and $1\rightarrow1$ transitions.
Scores are out of 24.}
\label{tab:disclosure-profile-changes}
\end{table}

Llama illustrates improvement accompanied by losses.
From Llama~2 to Llama~3.1, the score increased from 6
to 8 through four gains and two losses.
The gains concerned processing methods, purposes,
and techniques, together with compute hardware.
The losses concerned usage data used in training
and data-language composition.

Across the full Llama sequence, acquisition scores
declined from 2 to 1 to 0, while processing scores
changed from 3 to 0 to 3.
As Figure~\ref{fig:disclosure-overview} illustrates,
Llama~3.1 recovered processing disclosure without
restoring the named-source breakdown reported for
Llama~1. That source comparison illustrates a change
in documentary specificity; the acquisition scores
summarize all seven selected acquisition criteria.

\subsection{Persistent Gaps in Pretraining Disclosure}
\label{sec:persistent-disclosure-gaps}

\begin{figure*}[t]
    \centering
    \includegraphics[width=\textwidth]
        {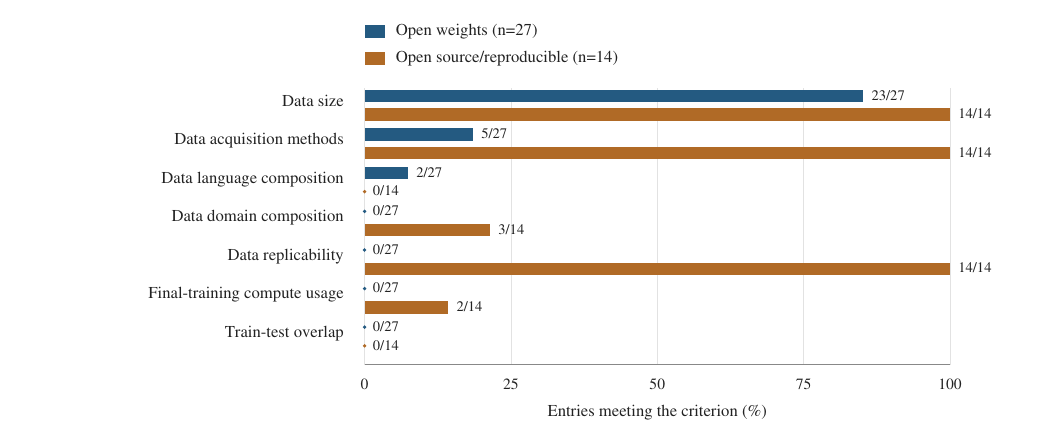}
    \caption{Selected disclosure criteria for open-weight
    ($n=27$) and open-source/reproducible ($n=14$) entries.
    Bars show percentages meeting each criterion;
    labels give counts. A zero denotes an unmet
    requirement.}
    \label{fig:persistent-disclosure-gaps}
\end{figure*}

Model stages, objectives, and data size were the most
consistently documented aspects of pretraining.
All 41 entries met the model-stages criterion,
and 37 met each of the objectives and data-size criteria.
Only two met language composition and three met domain
composition. None met external data access or
train--test overlap under the applied requirements.
Figure~\ref{fig:persistent-disclosure-gaps} compares
seven selected indicators across the groups.

Open-weight entries documented corpus size more often
than acquisition or composition.
Of the 27 entries, 23 met data size, five met acquisition
methods, one met public-dataset disclosure, and two
met language composition. None met domain composition,
data replicability, or training-code access.
Processing purpose was credited for 14 entries,
compared with ten for methods and six for techniques.

All 14 open-source/reproducible entries met the three
processing criteria, data replicability, and training-code
access. Nevertheless, none met language composition,
and only three met domain composition.
Broader artifact availability did not ensure satisfaction
of every selected disclosure requirement.

Compute reporting was also incomplete under the criteria:
only two entries met final-training compute usage,
and two met development duration.
Relevant quantitative information could be present
without meeting the complete requirement.
Some reports supplied GPU-hours or compute budgets
but lacked sufficient methodological or pretraining-stage
detail. Similarly, descriptions of decontamination
or overlap measurements for selected benchmarks
did not establish overlap across all relevant evaluations.
These results distinguish partial reporting from
satisfaction of the full coding requirements.

\section{Discussion}
\label{sec:discussion}

Our results suggest that pretraining disclosure is
a property of individual releases that needs to be
maintained as models develop. Later entries sometimes
provide less information under the selected criteria,
and improvements in total scores do not necessarily
preserve previously documented aspects of pretraining.
These findings have implications for maintaining
documentation and interpreting transparency assessments.

\subsection{Maintaining Disclosure across Releases}

\citet{xu2024tracing} show that transparency can evolve
differently across aspects of a model's lifecycle.
Our study extends this perspective across selected
families using a common set of pretraining criteria.
The comparisons identify where later releases add
information, retain earlier disclosures, or leave
previously documented aspects insufficiently specified.

This distinction matters when researchers consult
an earlier report to understand a later model.
A shared family name does not establish that the
same data sources, processing procedures, or training
arrangements remain applicable. At the same time,
developers need not repeat an unchanged description
in every report. An explicit reference can establish
which earlier documentation still describes the
later release.

Our coding recognized such references when the later
documentation explicitly established continuity.
The observed losses therefore concern requirements
not met by either the later disclosures or explicitly
applicable earlier documentation. They do not establish
that developers removed earlier documents, changed
their practices, or deliberately withheld information.
Because we examined disclosure available at assessment,
the comparisons also do not reconstruct documentation
as it existed on each release date.

We recommend that each release identify which
pretraining descriptions remain applicable, which
have changed, and where supporting evidence can be
found. A concise record linking each aspect of
pretraining to current or inherited documentation
would make continuity easier to verify. This would
also help distinguish an unchanged procedure from
an omission whose meaning cannot be determined
from the available evidence.

\subsection{What Disclosure Scores Establish}

The OLMo and Llama comparisons demonstrate that
a total score cannot establish continuity of information.
OLMo's stable score reflected gains offset by losses,
while Llama's improving score accompanied losses
in data-related criteria. These indicators support
different questions: hardware information helps
characterize training resources, whereas language
composition helps examine linguistic representation.
Additional information in one area does not answer
an unresolved question in another.

This finding complements
\citet{magnusson-etal-2023-reproducibility}, who
recommend tracking data-documentation practices
through specific checklist items. Our comparisons
show why this specificity also matters over time:
the same number of satisfied requirements can
correspond to different information across releases.

FMTI supplies explicit criteria for comparison,
but the resulting scores have a defined scope.
They measure whether evidence meets the selected
requirements; they do not directly measure
reproducibility, documentation usefulness, or
responsible development. As
\citet{liu-etal-2024-ecbd} argue in the context
of benchmark design, interpreting a measurement
requires clarity about what it supports.
We therefore use totals to summarize the extent
of disclosure and indicator profiles to establish
which information is documented.

The group comparison requires similar care.
Open-source/reproducible releases are expected
to provide broader access to development resources,
and some of these resources also receive credit
under the disclosure criteria. Their higher scores
therefore partly follow from the classification.
The informative questions are which requirements
remain unmet and whether disclosures persist
within subsequent releases.

\subsection{Implications for Independent Scrutiny}

The persistent gaps identify questions that researchers
cannot readily resolve from the reviewed documentation.
Corpus size does not establish the origins or
distribution of its contents.
\citet{baack2024commoncrawl} distinguish access
to crawl data from transparency about its subsequent
selection and filtering, while
\citet{dodge-etal-2021-documenting} demonstrate
the insights obtained by examining corpus construction.
Similarly, \citet{kreutzer-etal-2022-quality} show
that nominal multilingual coverage can obscure
substantial differences in data quality.

Our audit does not establish comparable problems
in the selected models' corpora. It identifies
where documentation does not meet the requirements
for describing relevant aspects of pretraining.
Available data and code may permit further investigation,
but deriving an answer from artifacts is different
from receiving an explicit, release-specific account.
This distinction helps explain why broader artifact
availability can accompany gaps in reported
corpus composition.

Train--test overlap illustrates the importance
of distinguishing partial disclosure from complete
reporting. \citet{zhang2025overlap} argue that
developers should publish training data or overlap
statistics to support interpretation of evaluations.
Their assessment recognizes reporting under different
conditions from our selected FMTI criterion.
Our zero scores therefore do not contradict evidence
that some developers publish useful overlap information.
They indicate that the reviewed evidence did not meet
the full requirement applied here.

Useful release documentation should specify the
pretraining data and procedures concerned, the
evaluations covered by overlap analyses, and the
limits of the reported information.
Together with explicit references to inherited
documentation, these details would support scrutiny
of individual releases and comparisons across them.
Our findings motivate assessing both how much
pretraining information is disclosed and whether
its applicability remains clear as models evolve.

\section{Conclusion}
\label{sec:conclusion}

We investigate how pretraining disclosure changes across
model releases through an audit of 41 entries from
17 families using 24 selected FMTI indicators.
Across 24 within-family comparisons, disclosure scores
decreased more often than they increased, and stable
or improving totals sometimes concealed losses in
particular information. These findings show that
later releases do not consistently preserve or expand
earlier disclosure. They motivate release-specific
documentation that makes changes and the applicability
of inherited information explicit, alongside assessments
that track individual disclosures as well as total scores.

\clearpage
\section*{Limitations}

\paragraph{Selected corpus.}
Our corpus is a selected sample of 41 entries from
17 families, not a representative survey of all language
models. Inclusion generally required a developer-authored
technical report, with official Marin retrospectives
treated as exceptions. This requirement may favor
better-documented releases and excludes models whose
disclosures appear only in other formats. The findings
therefore characterize the selected entries rather
than the prevalence of disclosure across the full
model ecosystem.

\paragraph{Comparability within families.}
The recorded sequences include architectural branches,
continued-pretraining releases, and grouped variants.
They are not controlled comparisons between otherwise
equivalent successors. Differences in model scope and
reporting conventions may contribute to score changes.
Families also contribute unequal numbers of comparisons,
so pooled transition counts give greater representation
to longer sequences.

\paragraph{Grouped variants.}
A grouped entry received credit when at least one
represented variant met a criterion. Different criteria
may therefore be supported by evidence concerning
different variants, and no single variant need satisfy
the group's complete indicator profile. This rule
captures disclosure available across the release group
but can obscure variation between model sizes.

\paragraph{Timing and inherited documentation.}
We evaluated information available at assessment,
rather than reconstructing disclosure on each original
release date. Earlier releases may have benefited
from subsequent documentation updates or a longer
period in which to accumulate supporting material.
Earlier documentation was credited only when its
applicability to a later release was explicitly
established. Unchanged practices without such a
statement could consequently remain uncredited.
Observed losses do not establish that information
was withdrawn or practices changed.

\paragraph{Measurement scope.}
The scores summarize 24 selected FMTI requirements
applied to pretraining. Binary coding does not capture
all differences in specificity, usefulness, or partial
disclosure. Equal weighting also does not imply that
the criteria are interchangeable or equally important
for every research question. An unmet requirement
may coexist with substantial relevant information,
and the scores do not directly establish reproducibility,
data quality, or responsible development.

\paragraph{Coding verification.}
AI-assisted coding was checked against supporting
sources by both authors, but verification was sequential
and based on existing records. The second reviewer
could therefore be influenced by earlier decisions.
Consensus resolution does not provide an independent
estimate of coding reliability, and missed evidence
or interpretive errors may remain.

\paragraph{Openness classification.}
Our openness groups are operational classifications,
not formal certifications or verified reproduction
outcomes. Training-code and training-data availability
inform both classification and some disclosure criteria.
The group difference is therefore partly structural
and should not be interpreted as an independent
effect of openness on disclosure.


\bibliography{custom}

\appendix

\section{Model Inventory}
\label{app:corpus}
Table~\ref{tab:appendix-inventory} lists the 41 entries included in our audit,
their family groupings, and associated technical documentation.
OW denotes open weights and OS denotes open source/reproducible, following
the operational definitions in Section~\ref{sec:model-selection}.
The order column identifies each entry's position in the selected family
sequence; it does not establish direct ancestry. Entries containing multiple
variants receive credit when at least one variant meets a criterion.

The search closed on 29 August 2026. We considered disclosure available for
each release at assessment, including earlier documentation whose continued
applicability was explicitly established. The listed references identify
technical reports and selected associated materials.

\clearpage
\onecolumn
\begingroup
\small
\setlength{\tabcolsep}{4pt}
\begin{longtable}{@{}p{1.6cm}p{4.1cm}rrp{6.5cm}@{}}
\caption{Selected entries and report references. Scores are out of 24.}
\label{tab:appendix-inventory}\\
\toprule
Family / group & Entry & Order & Score & Primary report reference(s)\\
\midrule
\endfirsthead
\multicolumn{5}{l}{\tablename\ \thetable\ (continued)}\\
\toprule
Family / group & Entry & Order & Score & Primary report reference(s)\\
\midrule
\endhead
\midrule
\multicolumn{5}{r}{\textit{Continued on next page}}\\
\endfoot
\bottomrule
\endlastfoot
Llama / OW & Llama 1 & 1 & 10 & \url{https://arxiv.org/abs/2302.13971}\\
Llama / OW & Llama 2 & 2 & 6 & \url{https://arxiv.org/abs/2307.09288}\\
Llama / OW & Llama 3.1 & 3 & 8 & \url{https://arxiv.org/abs/2407.21783}\\
Falcon / OW & Falcon 7B 40B 180B & 1 & 10 & \url{https://arxiv.org/abs/2311.16867}\newline \url{https://huggingface.co/datasets/tiiuae/falcon-refinedweb}\\
Falcon / OW & Falcon 2 11B & 2 & 4 & \url{https://arxiv.org/abs/2407.14885}\\
Falcon / OW & Falcon Mamba 7B & 3 & 7 & \url{https://arxiv.org/abs/2410.05355}\\
Falcon / OW & Falcon H1 & 4 & 6 & \url{https://arxiv.org/abs/2507.22448}\newline \url{https://github.com/tiiuae/falcon-h1}\\
Gemma / OW & Gemma 1 & 1 & 5 & \url{https://arxiv.org/abs/2403.08295}\\
Gemma / OW & Gemma 2 & 2 & 5 & \url{https://arxiv.org/abs/2408.00118}\\
Gemma / OW & Gemma 3 & 3 & 4 & \url{https://arxiv.org/abs/2503.19786}\\
Qwen / OW & Qwen 1 & 1 & 5 & \url{https://arxiv.org/abs/2309.16609}\\
Qwen / OW & Qwen 2 & 2 & 3 & \url{https://arxiv.org/abs/2407.10671}\\
Qwen / OW & Qwen 2.5 & 3 & 3 & \url{https://arxiv.org/abs/2412.15115}\\
Qwen / OW & Qwen 3 & 4 & 5 & \url{https://arxiv.org/pdf/2505.09388}\newline \url{https://chat.qwen.ai/legal-agreement/privacy-policy}\newline \url{https://qwenlm.github.io/blog/qwen3/}\newline \url{https://www.alibabacloud.com/help/en/model-studio/what-is-qwen-llm}\\
Phi / OW & Phi 1.5 & 1 & 5 & \url{https://arxiv.org/abs/2309.05463}\newline \url{https://arxiv.org/abs/2306.11644}\\
Phi / OW & Phi 3 & 2 & 9 & \url{https://arxiv.org/abs/2404.14219}\\
Phi / OW & Phi 4 & 3 & 7 & \url{https://arxiv.org/abs/2412.08905}\\
Phi / OW & Phi 4 mini multimodal & 4 & 3 & \url{https://arxiv.org/abs/2503.01743}\\
DeepSeek / OW & DeepSeek LLM 7B 67B & 1 & 6 & \url{https://arxiv.org/abs/2401.02954}\\
DeepSeek / OW & DeepSeekMoE 16B & 2 & 6 & \url{https://arxiv.org/abs/2401.06066}\newline \url{https://arxiv.org/abs/2401.02954}\\
DeepSeek / OW & DeepSeek-V2 V2-Lite & 3 & 6 & \url{https://arxiv.org/abs/2405.04434}\newline \url{https://arxiv.org/abs/2401.02954}\\
DeepSeek / OW & DeepSeek-V3 Base & 4 & 5 & \url{https://arxiv.org/abs/2412.19437}\newline \url{https://arxiv.org/abs/2501.12948}\\
DeepSeek / OW & DeepSeek-V3.2 Base & 5 & 4 & \url{https://arxiv.org/abs/2512.02556}\\
DeepSeek / OW & DeepSeek-V4-Flash & 6 & 3 & \url{https://arxiv.org/abs/2606.19348}\\
Mistral / OW & Mistral 7B family & 1 & 1 & \url{https://arxiv.org/abs/2310.06825}\\
Mistral / OW & Mixtral 8x7B & 2 & 1 & \url{https://arxiv.org/abs/2401.04088}\\
Mistral / OW & Ministral 3 & 3 & 3 & \url{https://arxiv.org/abs/2601.08584}\newline \url{https://mistral.ai/news/mistral-3}\\
Pythia / OS & Pythia Suite & 1 & 17 & \url{https://arxiv.org/abs/2304.01373}\\
OLMo / OS & OLMO 1 & 1 & 15 & \url{https://arxiv.org/abs/2402.00838}\\
OLMo / OS & OLMO 2 & 2 & 17 & \url{https://arxiv.org/abs/2501.00656}\\
OLMo / OS & OLMO 3 & 3 & 17 & \url{https://arxiv.org/abs/2512.13961}\\
Apertus / OS & Apertus & 1 & 15 & \url{https://arxiv.org/abs/2509.14233}\\
BLOOM / OS & BLOOM & 1 & 14 & \url{https://arxiv.org/abs/2211.05100}\\
Amber / OS & Amber 7B & 1 & 17 & \url{https://arxiv.org/abs/2312.06550}\\
K2 / OS & LLM 360 K2 & 1 & 17 & \url{https://arxiv.org/abs/2501.07124}\\
K2 / OS & LLM 360 K2 V2 & 2 & 15 & \url{https://arxiv.org/abs/2512.06201}\\
Marin / OS & Marin 8B & 1 & 16 & \url{https://marin.readthedocs.io/en/latest/reports/marin-8b-retro/}\\
Marin / OS & Marin 32B & 2 & 16 & \url{https://marin.readthedocs.io/en/latest/reports/marin-32b-retro/}\\
MAP-Neo / OS & MAP Neo 7B & 1 & 15 & \url{https://arxiv.org/abs/2405.19327}\\
GPT-NeoX / OS & GPT Neo X & 1 & 16 & \url{https://arxiv.org/abs/2204.06745}\\
Cerebras-GPT / OS & Cerebras & 1 & 18 & \url{https://arxiv.org/abs/2304.03208}\\
\end{longtable}
\endgroup
\twocolumn

\section{Pretraining Disclosure Criteria}
\label{app:indicators}
We selected 24 FMTI indicators addressing the origins and preparation of
pretraining data, corpus properties, training resources, and base-model
development. The subset was fixed before model coding. Below, we provide
the definitions and guidance retained from FMTI 2025
\citep{wan20252025,stanfordcrfm2025fmtireports}, followed by the reason
for including each criterion. Original identifiers are retained.

FMTI addresses the full development lifecycle. In our study, evidence for
these criteria was restricted to pretraining, including continued pretraining
within an included release. References to other stages in the original
guidance describe the broader framework and do not expand our assessment scope.

\subsection{Data Acquisition}
\paragraph{Data acquisition methods (ID 0).}

\textit{Definition.} What methods does the developer use to acquire data used to build the model?

\textit{FMTI guidance.} Which of the following data acquisition methods does the developer use: (i) acquiring existing public datasets, (ii) crawling the web, (iii) using data acquired via its existing products and services, (iv) licensing existing data from external parties, (v) having humans create or annotate new data, (vi) using models to generate new data, or (vii) other data acquisition methods not captured by the above. For example, if the developer uses reinforcement learning from human feedback to train models using model-generated outputs with human preference annotations, this would satisfy categories (v) and (vi). Alternatively, if the developer post-trains its model using off-the-shelf preference data (for example, the Alpaca dataset), this would satisfy category (i).

\textit{Relevance to pretraining.} Identifies the acquisition routes underlying the pretraining corpus.

\paragraph{Public datasets (ID 1).}

\textit{Definition.} What are the top-5 sources (by volume) of publicly available datasets acquired for building the model?

\textit{FMTI guidance.} We define a source as the entity or means by which the developer acquires data. We define the top-5 sources as the top-5 sources by data volume.

\textit{Relevance to pretraining.} Makes major public sources identifiable rather than leaving their origins implicit.

\paragraph{Crawling (ID 2).}

\textit{Definition.} If data collection involves web-crawling, what is the crawler name and opt-out protocol?

\textit{FMTI guidance.} We award this point for disclosure of the crawler name and opt-out protocols, including if/how they respect the Robots Exclusion Protocol (robots.txt).

\textit{Relevance to pretraining.} Documents direct collection and mechanisms by which websites can opt out.

\paragraph{Usage data used in training (ID 3).}

\textit{Definition.} What are the top-5 sources (by volume) of usage data from the developer's products and services that are used for building the model?

\textit{FMTI guidance.} We define usage data as data collected from the use of a developer's products or services.

\textit{Relevance to pretraining.} Identifies whether and where product or service usage contributes to pretraining.

\paragraph{Notice of usage data used in training (ID 4).}

\textit{Definition.} For the top-5 sources of usage data, how are users of these products and services made aware that this data is used for building the model?

\textit{FMTI guidance.} We define usage data notice as the proactive disclosure to users of how their data is used for model development. For example, via a pop-up with a description, a link to the privacy policy, or link to a description of company practices.

\textit{Relevance to pretraining.} Documents notice to users whose usage data enter pretraining.

\paragraph{Licensed data sources (ID 5).}

\textit{Definition.} What are the top-5 sources (by volume) of licensed data acquired for building the model?

\textit{FMTI guidance.} We define a source as the entity from which the developer acquires data. For example, the Associated Press is reportedly a source of licensed data for OpenAI.

\textit{Relevance to pretraining.} Identifies major externally licensed pretraining sources.

\paragraph{Licensed data compensation (ID 6).}

\textit{Definition.} For each of the top-5 sources of licensed data, are details related to compensation disclosed?

\textit{FMTI guidance.} We award this point if the model developer describes the compensation structure specified in the contract with the data source or indicates they are prohibited from sharing this information if contractually mandated.

\textit{Relevance to pretraining.} Documents compensation arrangements associated with licensed sources.

\subsection{Data Processing}
\paragraph{Data processing methods (ID 12).}

\textit{Definition.} What are the methods the developer uses to process acquired data to determine the data directly used in building the model?

\textit{FMTI guidance.} We will award this point for disclosure of all of the methods used to process acquired data. Data processing refers to any method that substantively changes the content of the data. For example, compression or changing the data file format is generally not in the scope of this indicator.

\textit{Relevance to pretraining.} Establishes the inventory of content-changing corpus preparation operations.

\paragraph{Data processing purpose (ID 13).}

\textit{Definition.} For each data processing method, what is its primary purpose?

\textit{FMTI guidance.} Data processing refers to any method that substantively changes the content of the data. For example, compression or changing the data file format is generally not in the scope of this indicator.

\textit{Relevance to pretraining.} Explains why each corpus preparation operation is applied.

\paragraph{Data processing techniques (ID 14).}

\textit{Definition.} For each data processing method, how does the developer implement the method?

\textit{FMTI guidance.} Data processing refers to any method that substantively changes the content of the data. For example, compression or changing the data file format is generally not in the scope of this indicator.

\textit{Relevance to pretraining.} Provides implementation information needed to understand the preparation recipe.

\subsection{Data Properties}
\paragraph{Data size (ID 15).}

\textit{Definition.} Is the size of the data used in building the model disclosed?

\textit{FMTI guidance.} To receive this point, the developer should report data size in appropriate units (e.g. bytes, words, tokens, images, frames) and broken down by modality. Data size should be reported to a precision of one significant figure (e.g. 4 trillion tokens, 200 thousand images). The size should reflect data directly used in building the model (i.e. training data) and not data that was acquired but unused, or data used to evaluate the model.

\textit{Relevance to pretraining.} Characterizes the quantity of data actually used in pretraining.

\paragraph{Data language composition (ID 16).}

\textit{Definition.} For all text data used in building the model, what is the composition of languages?

\textit{FMTI guidance.} To receive this point, the developer should report (i) all languages which make up at least 1\% of the data and their corresponding proportions and (ii) a brief description of how languages are labeled (if a publicly available tool is used, include a link to the tool). Proportions should be reported to a precision of two significant figures and should describe proportions of documents labeled with some langauge. An "Unknown" category may be included to denote documents where the language could not be identified.

\textit{Relevance to pretraining.} Characterizes linguistic representation and the method used to identify languages.

\paragraph{Data domain composition (ID 17).}

\textit{Definition.} For all the data used in building the model, what is the composition of domains covered in the data?

\textit{FMTI guidance.} To receive this point, the developer should report the composition of the main domains included in the data used to train the model. This data should be at a level of granularity lower than broad claims about training on "internet data". For example, this could include the proportion of data from e-commerce, social media, news, code, etc. based on the URLs from which the data is sourced. Proportions should be reported to a precision of one significant figure.

\textit{Relevance to pretraining.} Characterizes the distribution of subject areas or source domains.

\paragraph{External data access (ID 18).}

\textit{Definition.} Does a third-party have direct access to the data used to build the model?

\textit{FMTI guidance.} By a third-party, we mean entities that are financially independent of the developer. We will award this point if at least one such entity is named as having direct access to the data. With that said, we may award this point if the developer provides justifications for prohibiting access to narrowly-scoped parts of the data.

\textit{Relevance to pretraining.} Identifies independently named entities able to inspect the underlying data.

\paragraph{Data replicability (ID 19).}

\textit{Definition.} Is the data used to build the model described in enough detail to be externally replicable?

\textit{FMTI guidance.} We will award this point if the description contains (i) a list of all publicly available training data and where to obtain it and (ii) a list of all training data obtainable from third parties and where to obtain it. These conditions refer to criteria 2 and 3 under the OSI Open Source AI v1.0 definition.

\textit{Relevance to pretraining.} Tests whether acquisition information supports external reconstruction of the data.

\subsection{Compute}
\paragraph{Compute usage for final training run (ID 20).}

\textit{Definition.} Is the amount of compute used in the model's final training run disclosed?

\textit{FMTI guidance.} Compute should be reported in appropriate units, which most often will be floating point operations (FLOPs), along with a description of the measurement methodology, which may involve estimation. Compute should be reported to a precision of one significant figure (e.g. 5 x 10\textasciicircum{}25 FLOPs). This number should represent the compute used to train the final model across all model stages.

\textit{Relevance to pretraining.} Characterizes computational requirements and how they were measured or estimated.

\paragraph{Development duration for final training run (ID 22).}

\textit{Definition.} Is the amount of time required to build the model disclosed?

\textit{FMTI guidance.} The amount of time should be specified in terms of both the continuous duration of time required and the number of hardware hours used. The continuous duration of time required to build the model should be reported in weeks, days, or hours to a precision of one significant figure (e.g. 3 weeks). The number of hardware hours should be reported to a precision of one significant figure and include the type of hardware hours. No form of decomposition into phases of building the model is required for this indicator, but it should be clear what the duration refers to (e.g. training the model, or training and subsequent evaluation and red teaming).

\textit{Relevance to pretraining.} Characterizes elapsed time and hardware time for the relevant training run.

\paragraph{Compute hardware for final training run (ID 23).}

\textit{Definition.} For the primary hardware used to build the model, is the amount and type of hardware disclosed?

\textit{FMTI guidance.} In most cases, this indicator will be satisfied by information regarding the number and type of GPUs or TPUs used to train the model. The number of hardware units should be reported to a precision of one significant figure (e.g. 800 NVIDIA H100 GPUs). We will not award this point if (i) the training hardware generally used by the developer is disclosed, but the specific hardware for the given model is not, or (ii) the training hardware is disclosed, but the amount of hardware is not. We will award this point even if information about the interconnects between hardware units is not disclosed.

\textit{Relevance to pretraining.} Identifies the specific hardware quantity and type supporting pretraining.

\paragraph{Compute provider (ID 24).}

\textit{Definition.} Is the compute provider disclosed?

\textit{FMTI guidance.} For example, the compute provider may be the model developer in the case of a self-owned cluster, a cloud provider like Microsoft Azure, Google Cloud Platform, or Amazon Web Services, or a national supercomputer. In the event that compute is provided by multiple sources or is highly decentralized, we will award this point if a developer makes a reasonable effort to describe the distribution of hardware owners.

\textit{Relevance to pretraining.} Identifies the organization or infrastructure supplying training compute.

\subsection{Methods}
\paragraph{Model stages (ID 29).}

\textit{Definition.} Are all stages in the model development process disclosed?

\textit{FMTI guidance.} Stages refer to each identifiable step that constitutes a substantive change to the model during the model building process. We recognize that different developers may use different terminology for these stages, or conceptualize the stages differently. We will award this point if there is a clear and complete description of these stages.

\textit{Relevance to pretraining.} Establishes the stages included in the scoped pretraining process.

\paragraph{Model objectives (ID 30).}

\textit{Definition.} For all stages that are described, is there a clear description of the associated learning objectives or a clear characterization of the nature of this update to the model?

\textit{FMTI guidance.} We recognize that different developers may use different terminology for these stages, or conceptualize the stages differently. We will award this point if there is a clear description of the update to the model related to each stage, whether that is the intent of the stage (e.g. making the model less harmful), a mechanistic characterization (e.g. minimizing a specific loss function), or an empirical assessment (e.g. evaluation results conducted before and after the stage).

\textit{Relevance to pretraining.} Explains the learning objective or nature of updates in each included stage.

\paragraph{Code access (ID 31).}

\textit{Definition.} Does the developer release code that allows third-parties to train and run the model?

\textit{FMTI guidance.} The released code does not need to match the code used internally.

\textit{Relevance to pretraining.} Identifies code access supporting external training and execution.

\subsection{Model Dependencies}
\paragraph{Model dependencies (ID 36).}

\textit{Definition.} Is the model(s) the model is derived from disclosed?

\textit{FMTI guidance.} We will award this point for a comprehensive disclosure of the model or models on which the foundation model directly depends on or is derived from, as well as the method by which it was derived (e.g., through fine tuning, model merging, or distillation). Additionally, we will award a point if the developer discloses that the model is not dependent on or derived from any model.

\textit{Relevance to pretraining.} Establishes whether the base model was derived from an earlier model and how.

\subsection{Train--Test Overlap}
\paragraph{Train-test overlap (ID 45).}

\textit{Definition.} Does the developer measure and disclose the overlap between the training set and the dataset used to evaluate model capabilities?

\textit{FMTI guidance.} We will award this point if, with every capability evaluation for which the developer reports results, the developer reports the overlap between the training set of the model and the dataset used for evaluation, as well as the general methodology for computing train-test overlap (e.g. a description of how n-gram matching was used).

\textit{Relevance to pretraining.} Supports interpretation of evaluations in relation to possible pretraining exposure.

\section{Annotation Guidelines}
\label{app:protocol}
\subsection{AI-Assisted Coding}
\label{app:ai-assisted-coding}
ChatGPT assisted with understanding the FMTI criteria, extracting evidence,
and producing provisional coding records. December 2025 indicator definitions,
scores, and developer transparency reports were provided as in-context
reference material. Both authors subsequently verified the coding decisions.

\paragraph{Prompt for preparatory instruction.}
Its purpose was to understand the
FMTI reports before coding models

\begin{quote}
Carefully review the supplied December 2025 FMTI indicator definitions,
scores, and developer transparency reports. Explain how the reports apply
the criteria, use evidence, and justify scoring decisions. Identify the
requirements that should guide the next stage of the study.

Use clear, concise language while preserving evidence, qualifications,
and technical accuracy. Explain necessary technical terms and preserve
the authors' meaning and methodological detail when analyzing research.
\end{quote}

\paragraph{Coding instruction.}

\begin{quote}
Examine the provided model documentation against the 24 selected FMTI
indicators, definitions, and criterion notes. Use the published transparency
reports as examples of criterion application.

Consider only base-model pretraining and relevant continued pretraining.
Do not substitute evidence about instruction tuning, alignment, or downstream
fine-tuning. Carry earlier disclosures forward only when their applicability
to the later release is explicitly established.

For each indicator, record the relevant disclosure, source reference,
justification, and provisional binary score. Assign 1 when the requirement
is met and 0 otherwise. Do not infer undocumented practices or treat silence
as non-use. Identify uncertain interpretations for human review.
\end{quote}

\paragraph{Interpretation.}
The reported
results use the human-verified scores, whose evidence references and
justifications support examination of the final decisions.

\subsection{Evidence and Scoring Rules}
\paragraph{Evidence attribution.}
We used developer-authored technical reports and directly associated official
documentation. Evidence had to describe the included entry's pretraining
process. Descriptions of later instruction tuning or alignment did not
substitute for pretraining information. Where a statement combined stages
without identifying what applied to pretraining, it was insufficient to
support the criterion.

\paragraph{Inherited descriptions.}
We consulted earlier documentation when a later report explicitly stated
that the relevant data, procedures, or other details remained unchanged or
were inherited. Otherwise, we did not carry earlier disclosures forward.
This distinguishes documented continuity from an assumption based on a
shared model-family name.

\paragraph{Scoring.}
Each criterion received 1 when its requirements were met and 0 otherwise.
An explicit statement of non-use could receive credit under the corresponding
FMTI rule or precedent; silence did not establish non-use. A positive score
therefore indicates a qualifying disclosure, not necessarily use of a data
category. For grouped entries, evidence for different criteria may concern
different variants. The total counts disclosures available across the group.

\paragraph{Verification.}
ChatGPT assisted with provisional evidence extraction and coding using
published FMTI reports as in-context examples. For each decision, we retained
the disclosure, source reference, justification, and score. Both authors
checked every decision against the relevant sources, resolving discrepancies
through discussion. The second review examined existing records rather
than independently assigned labels.

\section{Examples of Criterion Application}
\label{app:examples}
The following examples illustrate the distinction between relevant information
and satisfaction of a complete disclosure requirement. Each example identifies
the evidence considered and the basis for the final score.

\subsection{Llama 1 / 1: Public datasets}

\textbf{Score:} 1.

\paragraph{Evidence.} The full seven-source mixture and shares are disclosed: CommonCrawl 67\%, C4 15\%, GitHub 4.5\%, Wikipedia 4.5\%, Books 4.5\%, arXiv 2.5\%, Stack Exchange 2\%.

\paragraph{Decision.} The largest public sources by volume are identifiable from the complete mixture.

\paragraph{Source.} \url{https://arxiv.org/abs/2302.13971}

\subsection{Llama 2 / 0: Data acquisition methods}

\textbf{Score:} 0.

\paragraph{Evidence.} The pretraining corpus is only described as a new mix of publicly available sources; existing public datasets versus direct crawling are not disambiguated and the acquisition-method inventory is not exhaustive.

\paragraph{Decision.} This is insufficient under the FMTI acquisition taxonomy.

\paragraph{Source.} \url{https://arxiv.org/abs/2307.09288}

\subsection{Llama 3.1 / 16: Data language composition}

\textbf{Score:} 0.

\paragraph{Evidence.} A fastText language-ID pipeline covering 176 languages is described, but whole-corpus language proportions for all languages at or above 1\% are not reported.

\paragraph{Decision.} FMTI requires proportions plus the labeling method.

\paragraph{Source.} \url{https://arxiv.org/abs/2407.21783}

\subsection{Llama 3.1 / 17: Data domain composition}

\textbf{Score:} 0.

\paragraph{Evidence.} The 50/25/17/8 mix is a broad training-category mixture (general knowledge, math/reasoning, code, multilingual), not a complete domain-composition analysis of the corpus.

\paragraph{Decision.} The Data domain composition criterion is not fully met.

\paragraph{Source.} \url{https://arxiv.org/abs/2407.21783}

\subsection{Llama 3.1 / 20: Compute usage for final training run}

\textbf{Score:} 0.

\paragraph{Evidence.} The report states a $3.8\times10^{25}$ FLOP pretraining budget, but does not provide an FMTI-style measurement/estimation methodology for the final run.

\paragraph{Decision.} The compute budget and GPU-hours provide relevant quantitative information, but do not establish the measurement or estimation methodology required by the criterion.

\paragraph{Source.} \url{https://arxiv.org/abs/2407.21783}

\subsection{Llama 3.1 / 45: Train-test overlap}

\textbf{Score:} 0.

\paragraph{Evidence.} The report provides an 8-gram contamination methodology and results for key benchmarks, but some results are unavailable/omitted and overlap is not reported for every capability evaluation.

\paragraph{Decision.} The every-evaluation requirement is not met.

\paragraph{Source.} \url{https://arxiv.org/abs/2407.21783}

\section{Supplementary Analyses}
\label{app:analysis}
Let $x_{ij}\in\{0,1\}$ denote entry $i$'s score on indicator $j$.
The overall score is $S_i=\sum_{j=1}^{24}x_{ij}$, and a subdomain score
is the sum over its selected indicators. For a group $\mathcal I$,
indicator prevalence is
\begin{equation}P_j(\mathcal I)=100|\mathcal I|^{-1}\sum_{i\in\mathcal I}x_{ij}.\end{equation}
Family-balanced prevalence assigns equal weight to each of the $F=17$ families:
\begin{equation}P_j^{(F)}=\frac{100}{F}\sum_{f=1}^{F}\frac{1}{n_f}\sum_{i\in f}x_{ij}.\end{equation}
Family coverage counts families with at least one entry receiving credit.
Neither statistic establishes prevalence in an unsampled model population.
Table~\ref{tab:appendix-prevalence} reports both entry-level and family-balanced summaries.

\begin{table*}[tp]
\centering
\small
\setlength{\tabcolsep}{4pt}
\caption{All selected indicators: entry counts, equal-family prevalence, and family coverage.}
\label{tab:appendix-prevalence}
\begin{tabular}{@{}rp{5.7cm}rrrrr@{}}
\toprule
ID & Indicator & OW & OS & All & Balanced (\%) & Families\\
\midrule
0 & Data acquisition methods & 5/27 & 14/14 & 19/41 & 66.7 & 13\\
1 & Public datasets & 1/27 & 8/14 & 9/41 & 37.3 & 7\\
2 & Crawling & 0/27 & 1/14 & 1/41 & 2.0 & 1\\
3 & Usage data used in training & 2/27 & 14/14 & 16/41 & 62.3 & 12\\
4 & Notice of usage data used in training & 1/27 & 14/14 & 15/41 & 60.3 & 11\\
5 & Licensed data sources & 1/27 & 13/14 & 14/41 & 54.4 & 10\\
6 & Licensed data compensation & 1/27 & 13/14 & 14/41 & 54.4 & 10\\
12 & Data processing methods & 10/27 & 14/14 & 24/41 & 73.0 & 15\\
13 & Data processing purpose & 14/27 & 14/14 & 28/41 & 80.4 & 16\\
14 & Data processing techniques & 6/27 & 14/14 & 20/41 & 68.6 & 13\\
15 & Data size & 23/27 & 14/14 & 37/41 & 92.2 & 16\\
16 & Data language composition & 2/27 & 0/14 & 2/41 & 3.4 & 2\\
17 & Data domain composition & 0/27 & 3/14 & 3/41 & 7.8 & 2\\
18 & External data access & 0/27 & 0/14 & 0/41 & 0.0 & 0\\
19 & Data replicability & 0/27 & 14/14 & 14/41 & 58.8 & 10\\
20 & Compute usage for final training run & 0/27 & 2/14 & 2/41 & 7.8 & 2\\
22 & Development duration for final training run & 1/27 & 1/14 & 2/41 & 7.8 & 2\\
23 & Compute hardware for final training run & 9/27 & 10/14 & 19/41 & 61.3 & 13\\
24 & Compute provider & 6/27 & 11/14 & 17/41 & 58.8 & 13\\
29 & Model stages & 27/27 & 14/14 & 41/41 & 100.0 & 17\\
30 & Model objectives & 23/27 & 14/14 & 37/41 & 92.2 & 17\\
31 & Code access & 0/27 & 14/14 & 14/41 & 58.8 & 10\\
36 & Model dependencies & 8/27 & 9/14 & 17/41 & 47.1 & 11\\
45 & Train-test overlap & 0/27 & 0/14 & 0/41 & 0.0 & 0\\
\bottomrule
\end{tabular}
\end{table*}
\clearpage

\paragraph{Sensitivity interpretation.}
Giving each family equal weight preserves the broad contrast between commonly
documented stages, objectives, and data size and the absence of credit for
external data access and train--test overlap. Rankings and ties can differ
between the two summaries; Table~\ref{tab:appendix-prevalence} reports the
complete values.

\paragraph{Release transitions.}
For adjacent entries $a,b$, gains count $x_{aj}=0,x_{bj}=1$, losses count
$x_{aj}=1,x_{bj}=0$, and retained disclosures count $x_{aj}=x_{bj}=1$.
Continued non-satisfaction counts $x_{aj}=x_{bj}=0$.
Thus $S_b-S_a=G_{ab}-L_{ab}$ and all four transition counts sum to 24.
Counts are aggregated across the selected comparisons, including repeated
transitions of an indicator. They are not normalized by family length or
subdomain size and do not constitute independent statistical trials.

\begin{table*}[tp]
\centering
\small
\setlength{\tabcolsep}{4pt}
\caption{All adjacent comparisons. G: gained; L: lost; R: retained; N: continued non-satisfaction.}
\label{tab:appendix-transitions}
\begin{tabular}{@{}p{1.5cm}p{3.8cm}p{3.8cm}rrrrr@{}}
\toprule
Family & Earlier entry & Later entry & Scores & G & L & R & N\\
\midrule
Llama & Llama 1 & Llama 2 & 10--6 & 3 & 7 & 3 & 11\\
Llama & Llama 2 & Llama 3.1 & 6--8 & 4 & 2 & 4 & 14\\
Falcon & Falcon 7B 40B 180B & Falcon 2 11B & 10--4 & 0 & 6 & 4 & 14\\
Falcon & Falcon 2 11B & Falcon Mamba 7B & 4--7 & 3 & 0 & 4 & 17\\
Falcon & Falcon Mamba 7B & Falcon H1 & 7--6 & 0 & 1 & 6 & 17\\
Gemma & Gemma 1 & Gemma 2 & 5--5 & 0 & 0 & 5 & 19\\
Gemma & Gemma 2 & Gemma 3 & 5--4 & 1 & 2 & 3 & 18\\
Qwen & Qwen 1 & Qwen 2 & 5--3 & 0 & 2 & 3 & 19\\
Qwen & Qwen 2 & Qwen 2.5 & 3--3 & 0 & 0 & 3 & 21\\
Qwen & Qwen 2.5 & Qwen 3 & 3--5 & 2 & 0 & 3 & 19\\
Phi & Phi 1.5 & Phi 3 & 5--9 & 5 & 1 & 4 & 14\\
Phi & Phi 3 & Phi 4 & 9--7 & 2 & 4 & 5 & 13\\
Phi & Phi 4 & Phi 4 mini multimodal & 7--3 & 0 & 4 & 3 & 17\\
DeepSeek & DeepSeek LLM 7B 67B & DeepSeekMoE 16B & 6--6 & 0 & 0 & 6 & 18\\
DeepSeek & DeepSeekMoE 16B & DeepSeek-V2 V2-Lite & 6--6 & 0 & 0 & 6 & 18\\
DeepSeek & DeepSeek-V2 V2-Lite & DeepSeek-V3 Base & 6--5 & 1 & 2 & 4 & 17\\
DeepSeek & DeepSeek-V3 Base & DeepSeek-V3.2 Base & 5--4 & 0 & 1 & 4 & 19\\
DeepSeek & DeepSeek-V3.2 Base & DeepSeek-V4-Flash & 4--3 & 0 & 1 & 3 & 20\\
Mistral & Mistral 7B family & Mixtral 8x7B & 1--1 & 0 & 0 & 1 & 23\\
Mistral & Mixtral 8x7B & Ministral 3 & 1--3 & 2 & 0 & 1 & 21\\
OLMo & OLMO 1 & OLMO 2 & 15--17 & 2 & 0 & 15 & 7\\
OLMo & OLMO 2 & OLMO 3 & 17--17 & 3 & 3 & 14 & 4\\
K2 & LLM 360 K2 & LLM 360 K2 V2 & 17--15 & 0 & 2 & 15 & 7\\
Marin & Marin 8B & Marin 32B & 16--16 & 0 & 0 & 16 & 8\\
\bottomrule
\end{tabular}
\end{table*}
\clearpage

\section{Complete Scores and Indicator Matrix}
\label{app:matrix}
Table~\ref{tab:appendix-scores} reports overall and subdomain scores for all
entries. Table~\ref{tab:appendix-matrix} provides their indicator profiles,
with columns corresponding to the original FMTI identifiers. A value of 1
indicates that the pretraining-scoped requirement was met.

\begin{table*}[tp]
\centering
\small
\setlength{\tabcolsep}{4pt}
\caption{Complete entry and subdomain scores. Denominators: 24, 7, 3, 5, 4, 3, 1, 1. (Part 1 of 2.)}
\label{tab:appendix-scores}
\begin{tabular}{@{}p{5.5cm}rrrrrrrr@{}}
\toprule
Entry & Total & Acq. & Proc. & Prop. & Comp. & Meth. & Dep. & Overlap\\
\midrule
Llama 1 & 10 & 2 & 3 & 1 & 2 & 2 & 0 & 0\\
Llama 2 & 6 & 1 & 0 & 2 & 1 & 2 & 0 & 0\\
Llama 3.1 & 8 & 0 & 3 & 1 & 2 & 2 & 0 & 0\\
Falcon 7B 40B 180B & 10 & 1 & 3 & 2 & 2 & 2 & 0 & 0\\
Falcon 2 11B & 4 & 0 & 0 & 1 & 1 & 2 & 0 & 0\\
Falcon Mamba 7B & 7 & 0 & 3 & 1 & 1 & 2 & 0 & 0\\
Falcon H1 & 6 & 0 & 3 & 1 & 0 & 2 & 0 & 0\\
Gemma 1 & 5 & 0 & 1 & 1 & 2 & 1 & 0 & 0\\
Gemma 2 & 5 & 0 & 1 & 1 & 2 & 1 & 0 & 0\\
Gemma 3 & 4 & 0 & 1 & 0 & 1 & 2 & 0 & 0\\
Qwen 1 & 5 & 0 & 2 & 1 & 0 & 2 & 0 & 0\\
Qwen 2 & 3 & 0 & 0 & 1 & 0 & 2 & 0 & 0\\
Qwen 2.5 & 3 & 0 & 0 & 1 & 0 & 2 & 0 & 0\\
Qwen 3 & 5 & 0 & 0 & 1 & 1 & 2 & 1 & 0\\
Phi 1.5 & 5 & 1 & 0 & 1 & 0 & 2 & 1 & 0\\
Phi 3 & 9 & 5 & 1 & 1 & 0 & 2 & 0 & 0\\
Phi 4 & 7 & 1 & 3 & 1 & 0 & 2 & 0 & 0\\
Phi 4 mini multimodal & 3 & 0 & 0 & 1 & 0 & 2 & 0 & 0\\
DeepSeek LLM 7B 67B & 6 & 0 & 2 & 1 & 0 & 2 & 1 & 0\\
DeepSeekMoE 16B & 6 & 0 & 2 & 1 & 0 & 2 & 1 & 0\\
DeepSeek-V2 V2-Lite & 6 & 0 & 2 & 1 & 0 & 2 & 1 & 0\\
DeepSeek-V3 Base & 5 & 0 & 0 & 1 & 1 & 2 & 1 & 0\\
\bottomrule
\end{tabular}
\end{table*}

\begin{table*}[tp]
\centering
\small
\setlength{\tabcolsep}{4pt}
\caption{Complete entry and subdomain scores. Denominators: 24, 7, 3, 5, 4, 3, 1, 1. (Part 2 of 2.)}
\label{tab:appendix-scores-2}
\begin{tabular}{@{}p{5.5cm}rrrrrrrr@{}}
\toprule
Entry & Total & Acq. & Proc. & Prop. & Comp. & Meth. & Dep. & Overlap\\
\midrule
DeepSeek-V3.2 Base & 4 & 0 & 0 & 1 & 0 & 2 & 1 & 0\\
DeepSeek-V4-Flash & 3 & 0 & 0 & 1 & 0 & 2 & 0 & 0\\
Mistral 7B family & 1 & 0 & 0 & 0 & 0 & 1 & 0 & 0\\
Mixtral 8x7B & 1 & 0 & 0 & 0 & 0 & 1 & 0 & 0\\
Ministral 3 & 3 & 0 & 0 & 0 & 0 & 2 & 1 & 0\\
Pythia Suite & 17 & 6 & 3 & 2 & 2 & 3 & 1 & 0\\
OLMO 1 & 15 & 6 & 3 & 2 & 1 & 3 & 0 & 0\\
OLMO 2 & 17 & 6 & 3 & 2 & 2 & 3 & 1 & 0\\
OLMO 3 & 17 & 7 & 3 & 3 & 1 & 3 & 0 & 0\\
Apertus & 15 & 6 & 3 & 2 & 1 & 3 & 0 & 0\\
BLOOM & 14 & 3 & 3 & 2 & 3 & 3 & 0 & 0\\
Amber 7B & 17 & 6 & 3 & 2 & 2 & 3 & 1 & 0\\
LLM 360 K2 & 17 & 5 & 3 & 3 & 2 & 3 & 1 & 0\\
LLM 360 K2 V2 & 15 & 5 & 3 & 3 & 0 & 3 & 1 & 0\\
Marin 8B & 16 & 5 & 3 & 2 & 2 & 3 & 1 & 0\\
Marin 32B & 16 & 5 & 3 & 2 & 2 & 3 & 1 & 0\\
MAP Neo 7B & 15 & 5 & 3 & 2 & 1 & 3 & 1 & 0\\
GPT Neo X & 16 & 6 & 3 & 2 & 2 & 3 & 0 & 0\\
Cerebras & 18 & 6 & 3 & 2 & 3 & 3 & 1 & 0\\
\bottomrule
\end{tabular}
\end{table*}
\clearpage

\begin{table*}[tp]
\centering
\small
\setlength{\tabcolsep}{4pt}
\caption{Final 41-entry binary indicator matrix. Indicators 0--16.}
\label{tab:appendix-matrix}
\begin{tabular}{@{}p{4.7cm}rrrrrrrrrrrr@{}}
\toprule
Entry & 0 & 1 & 2 & 3 & 4 & 5 & 6 & 12 & 13 & 14 & 15 & 16\\
\midrule
Llama 1 & 1 & 1 & 0 & 0 & 0 & 0 & 0 & 1 & 1 & 1 & 1 & 0\\
Llama 2 & 0 & 0 & 0 & 1 & 0 & 0 & 0 & 0 & 0 & 0 & 1 & 1\\
Llama 3.1 & 0 & 0 & 0 & 0 & 0 & 0 & 0 & 1 & 1 & 1 & 1 & 0\\
Falcon 7B 40B 180B & 1 & 0 & 0 & 0 & 0 & 0 & 0 & 1 & 1 & 1 & 1 & 1\\
Falcon 2 11B & 0 & 0 & 0 & 0 & 0 & 0 & 0 & 0 & 0 & 0 & 1 & 0\\
Falcon Mamba 7B & 0 & 0 & 0 & 0 & 0 & 0 & 0 & 1 & 1 & 1 & 1 & 0\\
Falcon H1 & 0 & 0 & 0 & 0 & 0 & 0 & 0 & 1 & 1 & 1 & 1 & 0\\
Gemma 1 & 0 & 0 & 0 & 0 & 0 & 0 & 0 & 0 & 1 & 0 & 1 & 0\\
Gemma 2 & 0 & 0 & 0 & 0 & 0 & 0 & 0 & 0 & 1 & 0 & 1 & 0\\
Gemma 3 & 0 & 0 & 0 & 0 & 0 & 0 & 0 & 0 & 1 & 0 & 0 & 0\\
Qwen 1 & 0 & 0 & 0 & 0 & 0 & 0 & 0 & 1 & 1 & 0 & 1 & 0\\
Qwen 2 & 0 & 0 & 0 & 0 & 0 & 0 & 0 & 0 & 0 & 0 & 1 & 0\\
Qwen 2.5 & 0 & 0 & 0 & 0 & 0 & 0 & 0 & 0 & 0 & 0 & 1 & 0\\
Qwen 3 & 0 & 0 & 0 & 0 & 0 & 0 & 0 & 0 & 0 & 0 & 1 & 0\\
Phi 1.5 & 1 & 0 & 0 & 0 & 0 & 0 & 0 & 0 & 0 & 0 & 1 & 0\\
Phi 3 & 1 & 0 & 0 & 1 & 1 & 1 & 1 & 0 & 1 & 0 & 1 & 0\\
Phi 4 & 1 & 0 & 0 & 0 & 0 & 0 & 0 & 1 & 1 & 1 & 1 & 0\\
Phi 4 mini multimodal & 0 & 0 & 0 & 0 & 0 & 0 & 0 & 0 & 0 & 0 & 1 & 0\\
DeepSeek LLM 7B 67B & 0 & 0 & 0 & 0 & 0 & 0 & 0 & 1 & 1 & 0 & 1 & 0\\
DeepSeekMoE 16B & 0 & 0 & 0 & 0 & 0 & 0 & 0 & 1 & 1 & 0 & 1 & 0\\
DeepSeek-V2 V2-Lite & 0 & 0 & 0 & 0 & 0 & 0 & 0 & 1 & 1 & 0 & 1 & 0\\
DeepSeek-V3 Base & 0 & 0 & 0 & 0 & 0 & 0 & 0 & 0 & 0 & 0 & 1 & 0\\
DeepSeek-V3.2 Base & 0 & 0 & 0 & 0 & 0 & 0 & 0 & 0 & 0 & 0 & 1 & 0\\
DeepSeek-V4-Flash & 0 & 0 & 0 & 0 & 0 & 0 & 0 & 0 & 0 & 0 & 1 & 0\\
Mistral 7B family & 0 & 0 & 0 & 0 & 0 & 0 & 0 & 0 & 0 & 0 & 0 & 0\\
Mixtral 8x7B & 0 & 0 & 0 & 0 & 0 & 0 & 0 & 0 & 0 & 0 & 0 & 0\\
Ministral 3 & 0 & 0 & 0 & 0 & 0 & 0 & 0 & 0 & 0 & 0 & 0 & 0\\
Pythia Suite & 1 & 1 & 0 & 1 & 1 & 1 & 1 & 1 & 1 & 1 & 1 & 0\\
OLMO 1 & 1 & 1 & 0 & 1 & 1 & 1 & 1 & 1 & 1 & 1 & 1 & 0\\
OLMO 2 & 1 & 1 & 0 & 1 & 1 & 1 & 1 & 1 & 1 & 1 & 1 & 0\\
OLMO 3 & 1 & 1 & 1 & 1 & 1 & 1 & 1 & 1 & 1 & 1 & 1 & 0\\
Apertus & 1 & 1 & 0 & 1 & 1 & 1 & 1 & 1 & 1 & 1 & 1 & 0\\
BLOOM & 1 & 0 & 0 & 1 & 1 & 0 & 0 & 1 & 1 & 1 & 1 & 0\\
Amber 7B & 1 & 1 & 0 & 1 & 1 & 1 & 1 & 1 & 1 & 1 & 1 & 0\\
LLM 360 K2 & 1 & 0 & 0 & 1 & 1 & 1 & 1 & 1 & 1 & 1 & 1 & 0\\
LLM 360 K2 V2 & 1 & 0 & 0 & 1 & 1 & 1 & 1 & 1 & 1 & 1 & 1 & 0\\
Marin 8B & 1 & 0 & 0 & 1 & 1 & 1 & 1 & 1 & 1 & 1 & 1 & 0\\
Marin 32B & 1 & 0 & 0 & 1 & 1 & 1 & 1 & 1 & 1 & 1 & 1 & 0\\
MAP Neo 7B & 1 & 0 & 0 & 1 & 1 & 1 & 1 & 1 & 1 & 1 & 1 & 0\\
GPT Neo X & 1 & 1 & 0 & 1 & 1 & 1 & 1 & 1 & 1 & 1 & 1 & 0\\
Cerebras & 1 & 1 & 0 & 1 & 1 & 1 & 1 & 1 & 1 & 1 & 1 & 0\\
\bottomrule
\end{tabular}
\end{table*}

\begin{table*}[tp]
\centering
\small
\setlength{\tabcolsep}{4pt}
\caption{Final 41-entry binary indicator matrix. Indicators 17--45.}
\label{tab:appendix-matrix-continued}
\begin{tabular}{@{}p{4.7cm}rrrrrrrrrrrr@{}}
\toprule
Entry & 17 & 18 & 19 & 20 & 22 & 23 & 24 & 29 & 30 & 31 & 36 & 45\\
\midrule
Llama 1 & 0 & 0 & 0 & 0 & 1 & 1 & 0 & 1 & 1 & 0 & 0 & 0\\
Llama 2 & 0 & 0 & 0 & 0 & 0 & 0 & 1 & 1 & 1 & 0 & 0 & 0\\
Llama 3.1 & 0 & 0 & 0 & 0 & 0 & 1 & 1 & 1 & 1 & 0 & 0 & 0\\
Falcon 7B 40B 180B & 0 & 0 & 0 & 0 & 0 & 1 & 1 & 1 & 1 & 0 & 0 & 0\\
Falcon 2 11B & 0 & 0 & 0 & 0 & 0 & 1 & 0 & 1 & 1 & 0 & 0 & 0\\
Falcon Mamba 7B & 0 & 0 & 0 & 0 & 0 & 1 & 0 & 1 & 1 & 0 & 0 & 0\\
Falcon H1 & 0 & 0 & 0 & 0 & 0 & 0 & 0 & 1 & 1 & 0 & 0 & 0\\
Gemma 1 & 0 & 0 & 0 & 0 & 0 & 1 & 1 & 1 & 0 & 0 & 0 & 0\\
Gemma 2 & 0 & 0 & 0 & 0 & 0 & 1 & 1 & 1 & 0 & 0 & 0 & 0\\
Gemma 3 & 0 & 0 & 0 & 0 & 0 & 1 & 0 & 1 & 1 & 0 & 0 & 0\\
Qwen 1 & 0 & 0 & 0 & 0 & 0 & 0 & 0 & 1 & 1 & 0 & 0 & 0\\
Qwen 2 & 0 & 0 & 0 & 0 & 0 & 0 & 0 & 1 & 1 & 0 & 0 & 0\\
Qwen 2.5 & 0 & 0 & 0 & 0 & 0 & 0 & 0 & 1 & 1 & 0 & 0 & 0\\
Qwen 3 & 0 & 0 & 0 & 0 & 0 & 0 & 1 & 1 & 1 & 0 & 1 & 0\\
Phi 1.5 & 0 & 0 & 0 & 0 & 0 & 0 & 0 & 1 & 1 & 0 & 1 & 0\\
Phi 3 & 0 & 0 & 0 & 0 & 0 & 0 & 0 & 1 & 1 & 0 & 0 & 0\\
Phi 4 & 0 & 0 & 0 & 0 & 0 & 0 & 0 & 1 & 1 & 0 & 0 & 0\\
Phi 4 mini multimodal & 0 & 0 & 0 & 0 & 0 & 0 & 0 & 1 & 1 & 0 & 0 & 0\\
DeepSeek LLM 7B 67B & 0 & 0 & 0 & 0 & 0 & 0 & 0 & 1 & 1 & 0 & 1 & 0\\
DeepSeekMoE 16B & 0 & 0 & 0 & 0 & 0 & 0 & 0 & 1 & 1 & 0 & 1 & 0\\
DeepSeek-V2 V2-Lite & 0 & 0 & 0 & 0 & 0 & 0 & 0 & 1 & 1 & 0 & 1 & 0\\
DeepSeek-V3 Base & 0 & 0 & 0 & 0 & 0 & 1 & 0 & 1 & 1 & 0 & 1 & 0\\
DeepSeek-V3.2 Base & 0 & 0 & 0 & 0 & 0 & 0 & 0 & 1 & 1 & 0 & 1 & 0\\
DeepSeek-V4-Flash & 0 & 0 & 0 & 0 & 0 & 0 & 0 & 1 & 1 & 0 & 0 & 0\\
Mistral 7B family & 0 & 0 & 0 & 0 & 0 & 0 & 0 & 1 & 0 & 0 & 0 & 0\\
Mixtral 8x7B & 0 & 0 & 0 & 0 & 0 & 0 & 0 & 1 & 0 & 0 & 0 & 0\\
Ministral 3 & 0 & 0 & 0 & 0 & 0 & 0 & 0 & 1 & 1 & 0 & 1 & 0\\
Pythia Suite & 0 & 0 & 1 & 0 & 0 & 1 & 1 & 1 & 1 & 1 & 1 & 0\\
OLMO 1 & 0 & 0 & 1 & 0 & 0 & 0 & 1 & 1 & 1 & 1 & 0 & 0\\
OLMO 2 & 0 & 0 & 1 & 1 & 0 & 0 & 1 & 1 & 1 & 1 & 1 & 0\\
OLMO 3 & 1 & 0 & 1 & 0 & 0 & 1 & 0 & 1 & 1 & 1 & 0 & 0\\
Apertus & 0 & 0 & 1 & 0 & 0 & 0 & 1 & 1 & 1 & 1 & 0 & 0\\
BLOOM & 0 & 0 & 1 & 0 & 1 & 1 & 1 & 1 & 1 & 1 & 0 & 0\\
Amber 7B & 0 & 0 & 1 & 0 & 0 & 1 & 1 & 1 & 1 & 1 & 1 & 0\\
LLM 360 K2 & 1 & 0 & 1 & 0 & 0 & 1 & 1 & 1 & 1 & 1 & 1 & 0\\
LLM 360 K2 V2 & 1 & 0 & 1 & 0 & 0 & 0 & 0 & 1 & 1 & 1 & 1 & 0\\
Marin 8B & 0 & 0 & 1 & 0 & 0 & 1 & 1 & 1 & 1 & 1 & 1 & 0\\
Marin 32B & 0 & 0 & 1 & 0 & 0 & 1 & 1 & 1 & 1 & 1 & 1 & 0\\
MAP Neo 7B & 0 & 0 & 1 & 0 & 0 & 1 & 0 & 1 & 1 & 1 & 1 & 0\\
GPT Neo X & 0 & 0 & 1 & 0 & 0 & 1 & 1 & 1 & 1 & 1 & 0 & 0\\
Cerebras & 0 & 0 & 1 & 1 & 0 & 1 & 1 & 1 & 1 & 1 & 1 & 0\\
\bottomrule
\end{tabular}
\end{table*}
\clearpage

\section{Disclosure Relationships and Changed Criteria}
\label{app:disclosure-relations}

\subsection{Information Accompanying Broader Disclosures}
Table~\ref{tab:focused-disclosure-relations} examines whether entries meeting
one requirement also meet a related requirement. The arrow denotes a conditional
description, not causation or a logical implication. For each row, the denominator
counts entries satisfying the first condition and the numerator counts those
satisfying both. ``Language or domain'' and ``hardware or provider'' each use
the union of the two criteria, counting an entry once even when it meets both.

Only 5 of the 37 entries credited for data size also met language or domain
composition requirements. Similarly, only 2 of 24 entries credited for hardware
or provider disclosure met compute usage. Broad quantitative descriptions
therefore did not consistently accompany the more specific information captured
by these criteria.

Processing disclosures were more closely associated: 24 of 28 entries credited
for purpose also met methods, and 20 of those 24 met techniques. The pooled
usage-data-to-notice result was 15/16, but the open-weight result was only 1/2;
the small denominator limits interpretation. Explicit non-use can receive credit
under the coding rules, so these relationships concern qualifying disclosures
rather than proving that usage data were collected. All code-access entries
belonged to the open-source/reproducible group; the pooled dependency result
therefore describes that group alone. These are descriptive associations between
binary criteria, not evidence of independent effects of openness.

\begin{table*}[tp]
\centering
\small
\setlength{\tabcolsep}{4pt}
\caption{Selected disclosure relationships. Each cell reports entries meeting
both conditions divided by entries meeting the first condition.}
\label{tab:focused-disclosure-relations}
\begin{tabular}{@{}p{5.5cm}p{2.8cm}p{2.8cm}p{3.0cm}@{}}
\toprule
Relationship & All entries & Open weights & \shortstack[l]{Open source/\\reproducible}\\
\midrule
\multicolumn{4}{l}{\textit{Details accompanying broader disclosures}}\\
Data size $\rightarrow$ language or domain composition & 5/37 (13.5\%) & 2/23 (8.7\%) & 3/14 (21.4\%)\\
Hardware or provider $\rightarrow$ compute usage & 2/24 (8.3\%) & 0/11 (0.0\%) & 2/13 (15.4\%)\\
\addlinespace
\multicolumn{4}{l}{\textit{Related disclosure criteria}}\\
Processing purpose $\rightarrow$ processing methods & 24/28 (85.7\%) & 10/14 (71.4\%) & 14/14 (100.0\%)\\
Processing methods $\rightarrow$ processing techniques & 20/24 (83.3\%) & 6/10 (60.0\%) & 14/14 (100.0\%)\\
Usage data $\rightarrow$ notice & 15/16 (93.8\%) & 1/2 (50.0\%) & 14/14 (100.0\%)\\
Code access $\rightarrow$ model dependencies & 9/14 (64.3\%) & n.a. & 9/14 (64.3\%)\\
\bottomrule
\end{tabular}
\par\smallskip
\begin{minipage}{\textwidth}\footnotesize
n.a.: no open-weight entry met code access, so the conditional percentage is
undefined. The pooled code-access result comes entirely from the
open-source/reproducible group. Percentages are descriptive.
\end{minipage}
\end{table*}

\subsection{Criterion-Level Release Transitions}
Table~\ref{tab:full_transitions_landscape} extends the transition summary with
the identities of the gained and lost criteria. Original FMTI identifiers
refer to the definitions in Appendix~\ref{app:indicators}. The four transition
counts sum to 24 for each comparison, and the score difference equals gains
minus losses. The 24 comparisons comprise 20 open-weight and four
open-source/reproducible comparisons.

The Llama~2 to Llama~3.1 comparison gains the three processing criteria
(12--14) and hardware (23), while losing usage-data disclosure (3) and language
composition (16). OLMo~2 to OLMo~3 gains crawling (2), domain composition (17),
and hardware (23), offset by losses in compute usage (20), provider (24),
and dependencies (36). The full table also distinguishes stable totals with
no indicator changes from stable totals with replacement, as in OLMo.
No changed criteria are recorded for Gemma~1 to Gemma~2, Qwen~2 to Qwen~2.5,
the first two DeepSeek comparisons, Mistral~7B to Mixtral, or the Marin pair.
Losses identify requirements unmet by applicable evidence for the later entry;
they do not establish withdrawal of information or direct model ancestry.

\begin{table*}[tp]
\centering
\small
\setlength{\tabcolsep}{3pt}
\caption{All ordered release comparisons and changed criteria. G: gained
($0\to1$); L: lost ($1\to0$); N: still unmet ($0\to0$); R: retained
($1\to1$). Changed criteria use original FMTI IDs; -- denotes none.
MM abbreviates multimodal.}
\label{tab:full_transitions_landscape}
\begin{tabular}{@{}p{5.0cm}r r r r r r p{2.0cm}p{2.2cm}@{}}
\toprule
Comparison & Scores & $\Delta$ & G & L & N & R & Gained IDs & Lost IDs\\
\midrule
Llama 1 $\to$ Llama 2 & $10\to 6$ & -4 & 3 & 7 & 11 & 3 & 3, 16, 24 & 0, 1, 12, 13, 14, 22, 23\\
Llama 2 $\to$ Llama 3.1 & $6\to 8$ & +2 & 4 & 2 & 14 & 4 & 12, 13, 14, 23 & 3, 16\\
Falcon 7B 40B 180B $\to$ Falcon 2 11B & $10\to 4$ & -6 & 0 & 6 & 14 & 4 & -- & 0, 12, 13, 14, 16, 24\\
Falcon 2 11B $\to$ Falcon Mamba 7B & $4\to 7$ & +3 & 3 & 0 & 17 & 4 & 12, 13, 14 & --\\
Falcon Mamba 7B $\to$ Falcon H1 & $7\to 6$ & -1 & 0 & 1 & 17 & 6 & -- & 23\\
Gemma 1 $\to$ Gemma 2 & $5\to 5$ & +0 & 0 & 0 & 19 & 5 & -- & --\\
Gemma 2 $\to$ Gemma 3 & $5\to 4$ & -1 & 1 & 2 & 18 & 3 & 30 & 15, 24\\
Qwen 1 $\to$ Qwen 2 & $5\to 3$ & -2 & 0 & 2 & 19 & 3 & -- & 12, 13\\
Qwen 2 $\to$ Qwen 2.5 & $3\to 3$ & +0 & 0 & 0 & 21 & 3 & -- & --\\
Qwen 2.5 $\to$ Qwen 3 & $3\to 5$ & +2 & 2 & 0 & 19 & 3 & 24, 36 & --\\
Phi 1.5 $\to$ Phi 3 & $5\to 9$ & +4 & 5 & 1 & 14 & 4 & 3, 4, 5, 6, 13 & 36\\
Phi 3 $\to$ Phi 4 & $9\to 7$ & -2 & 2 & 4 & 13 & 5 & 12, 14 & 3, 4, 5, 6\\
Phi 4 $\to$ Phi 4 mini MM & $7\to 3$ & -4 & 0 & 4 & 17 & 3 & -- & 0, 12, 13, 14\\
DeepSeek\allowbreak  LLM 7B 67B $\to$ DeepSeek\allowbreak MoE 16B & $6\to 6$ & +0 & 0 & 0 & 18 & 6 & -- & --\\
DeepSeek\allowbreak MoE 16B $\to$ DeepSeek\allowbreak -V2 V2-Lite & $6\to 6$ & +0 & 0 & 0 & 18 & 6 & -- & --\\
DeepSeek\allowbreak -V2 V2-Lite $\to$ DeepSeek\allowbreak -V3 Base & $6\to 5$ & -1 & 1 & 2 & 17 & 4 & 23 & 12, 13\\
DeepSeek\allowbreak -V3 Base $\to$ DeepSeek\allowbreak -V3.2 Base & $5\to 4$ & -1 & 0 & 1 & 19 & 4 & -- & 23\\
DeepSeek\allowbreak -V3.2 Base $\to$ DeepSeek\allowbreak -V4-Flash & $4\to 3$ & -1 & 0 & 1 & 20 & 3 & -- & 36\\
Mistral 7B family $\to$ Mixtral 8x7B & $1\to 1$ & +0 & 0 & 0 & 23 & 1 & -- & --\\
Mixtral 8x7B $\to$ Ministral 3 & $1\to 3$ & +2 & 2 & 0 & 21 & 1 & 30, 36 & --\\
OLMO 1 $\to$ OLMO 2 & $15\to 17$ & +2 & 2 & 0 & 7 & 15 & 20, 36 & --\\
OLMO 2 $\to$ OLMO 3 & $17\to 17$ & +0 & 3 & 3 & 4 & 14 & 2, 17, 23 & 20, 24, 36\\
LLM 360 K2 $\to$ LLM 360 K2 V2 & $17\to 15$ & -2 & 0 & 2 & 7 & 15 & -- & 23, 24\\
Marin 8B $\to$ Marin 32B & $16\to 16$ & +0 & 0 & 0 & 8 & 16 & -- & --\\
\bottomrule
\end{tabular}
\end{table*}
\clearpage

\section{Literature Search}
\label{app:scoping}
An exploratory search beginning on 14 May 2026 informed the scope of the
study. We used Google Scholar Labs to identify work on pretraining provenance,
corpus construction, filtering, reproducibility, openness, and changes in
disclosure. The search also covered documentation of major corpora, including
Common Crawl, C4, The Pile, CC-100, RefinedWeb, and RedPajama.
The resulting literature collection contained 51 distinct studies.
This search supported methodological scoping and the related-work discussion;
it was not intended as a systematic review.

\end{document}